\pdfoutput=1
\documentclass{article}
\PassOptionsToPackage{numbers,sort&compress}{natbib}
\usepackage[dblblindworkshop,final]{neurips_2026}
\workshoptitle{Foundation Models for the Brain and Body}
\usepackage[utf8]{inputenc}
\usepackage[T1]{fontenc}
\usepackage{hyperref}
\usepackage{url}
\usepackage{booktabs}
\usepackage{graphicx}
\usepackage{adjustbox}
\usepackage{amsmath,amsfonts}
\usepackage{enumitem}
\usepackage{wrapfig}
\usepackage{microtype}
\usepackage{xcolor}
\usepackage{multirow}
\usepackage{makecell}
\usepackage[most]{tcolorbox}
\newtcolorbox{eegresponse}[1][]{
  enhanced,breakable,width=0.90\textwidth,center,
  colback=gray!5,colframe=gray!40,boxrule=0.6pt,arc=2mm,
  left=4pt,right=4pt,top=4pt,bottom=4pt,
  fontupper=\ttfamily\tiny\raggedright,title={#1}}
\newcommand{\ourmethod}{BraVista}
\title{Multimodal LLMs Can Learn to Read Brain Signals: A Vision--Language Model for Unified Multi-Task EEG Decoding}
\author{%
  Parastoo Azizeddin$^{\dagger 1}$ \quad Omid Sharafi$^{\dagger 1}$ \quad Maryam M. Shanechi$^{*1\text{--}4}$ \\
  \normalfont University of Southern California, Los Angeles, CA \\
  \normalfont\texttt{\{azizeddi,osharafi,shanechi\}@usc.edu}
}
\begin{document}
\maketitle
\begingroup
\renewcommand{\thefootnote}{$*$}\footnotetext{Corresponding author: \texttt{shanechi@usc.edu}.}
\renewcommand{\thefootnote}{$\dagger$}\footnotetext{Equal contribution.}
\renewcommand{\thefootnote}{1}\footnotetext{Ming Hsieh Department of Electrical and Computer Engineering, University of Southern California}
\renewcommand{\thefootnote}{2}\footnotetext{Thomas Lord Department of Computer Science, University of Southern California}
\renewcommand{\thefootnote}{3}\footnotetext{Alfred E. Mann Department of Biomedical Engineering, University of Southern California}
\renewcommand{\thefootnote}{4}\footnotetext{Neuroscience Graduate Program, University of Southern California}
\endgroup
\begin{abstract}

Learning EEG representations that generalize across cognitive tasks, subjects, and recording conditions remains a key challenge in electroencephalography (EEG) decoding. 
Recent advances in foundation models have improved EEG decoding performance, yet a fundamental open question remains: how to effectively interface neural signals with these models to enable multi-task learning across datasets. 
To investigate this question, we introduce \textbf{BraVista}, a visual--language framework that encodes multichannel EEG signals as structured images and enables multi-task learning through instruction-conditioned vision--language models (VLMs). Our approach relies on continued post-training of a general-domain VLM, leveraging its visual and linguistic priors to adapt to neural signals without a separate large-scale EEG-specific pretraining stage.
We evaluate \ourmethod{} on four datasets spanning sleep staging, emotion recognition, cognitive workload classification, and abnormal EEG detection, showing strong performance across these tasks. Further analyses show that the choice of EEG-to-image representation is critical to performance. Moreover, through controlled perturbations of the EEG signal, we observe a gradual performance degradation under increasing noise, suggesting that the model relies on EEG-relevant information rather than superficial visual patterns.
Together, these findings establish structured visual representations as an effective and scalable interface between neural signals and general-domain foundation models for unified multi-task EEG decoding.
 
\end{abstract}
\section{Introduction}
Electroencephalography (EEG) supports diverse cognitive and clinical applications, including epilepsy diagnosis~\cite{ahmad2023hybrid}, sleep staging~\cite{yang2021self}, affective computing~\cite{jenke2014feature}, cognitive-state decoding, and abnormality detection. However, EEG signals are highly variable across subjects, recording protocols, and acquisition systems~\cite{lotte2018review}. This heterogeneity makes it difficult to learn representations that can be shared across datasets and decoding tasks. Conventional deep learning approaches, including convolutional and recurrent networks~\cite{schirrmeister2017deep,lawhern2018eegnet,wang2018lstm,roy2019deep,phan2019seqsleepnet,ding2022tsception}, CNN--LSTM models~\cite{zhang2019novel,wang20232d}, and CNN--Transformer architectures~\cite{song2022eeg,phan2022sleeptransformer}, are typically trained for a single dataset or task. Recent EEG foundation models address this challenge by pretraining on large EEG corpora to learn transferable representations~\cite{kostas2021bendr,yi2023learning,yang2023biot,jiang2024large,wang2024cbramod,wang2024eegpt,ouahidi2025reve}, with related advances in representation learning for other neural recording modalities~\cite{oganesian2025barista,erturk2025crossmodal,javadzadeh2025functional,hosseini2026crosssubject}. 
However, they require a large-scale EEG-specific pretraining stage followed by separate adaptation for each downstream task, leaving unified multi-task decoding within a single model underexplored.

Recent advances in large language models (LLMs) enable unified multi-task prediction by allowing a single model to adapt to diverse tasks through instruction conditioning~\cite{brown2020language,touvron2023llama}. Existing EEG-language models couple EEG features or learned neural tokenizers with language models, enabling EEG tasks to be formulated as instruction-conditioned prediction problems~\cite{kim2024eeg,jiang2024neurolm,camaret2025eeg,lu2025unimind,zeng2025wavemind}. However, they rely on EEG-specific pretraining to align neural signals with language, making performance dependent on the scale and diversity of the EEG pretraining corpus~\cite{jiang2025data}. We ask whether a general-domain vision--language model (VLM)~\cite{liu2023visual} can instead be interfaced with EEG through a representation it already knows how to process: an image. This is motivated by image-based time-series modeling~\cite{jin2023time,zhong2025time,zhang2025timemaster,khezresmaeilzadeh2025vista,he2026harnessing}, but its effectiveness for unified EEG decoding remains unclear.

We introduce \textbf{BraVista}, which arranges channel-wise short-time Fourier transform (STFT) spectrograms into one structured image and pairs it with a natural-language task instruction. Via continued post-training, we adapt a Qwen3-VL-2B model to EEG datasets. Our study emphasizes three findings: (1) a single BraVista model achieves strong decoding across four distinct BCI tasks; (2) STFT images provide a substantially better VLM interface than raw-signal plots or topographies; and (3) controlled corruption produces gradual performance loss, indicating reliance on EEG signal structure rather than superficial visual cues.

\begin{figure*}[!h]
\centering
\includegraphics[width=\textwidth]{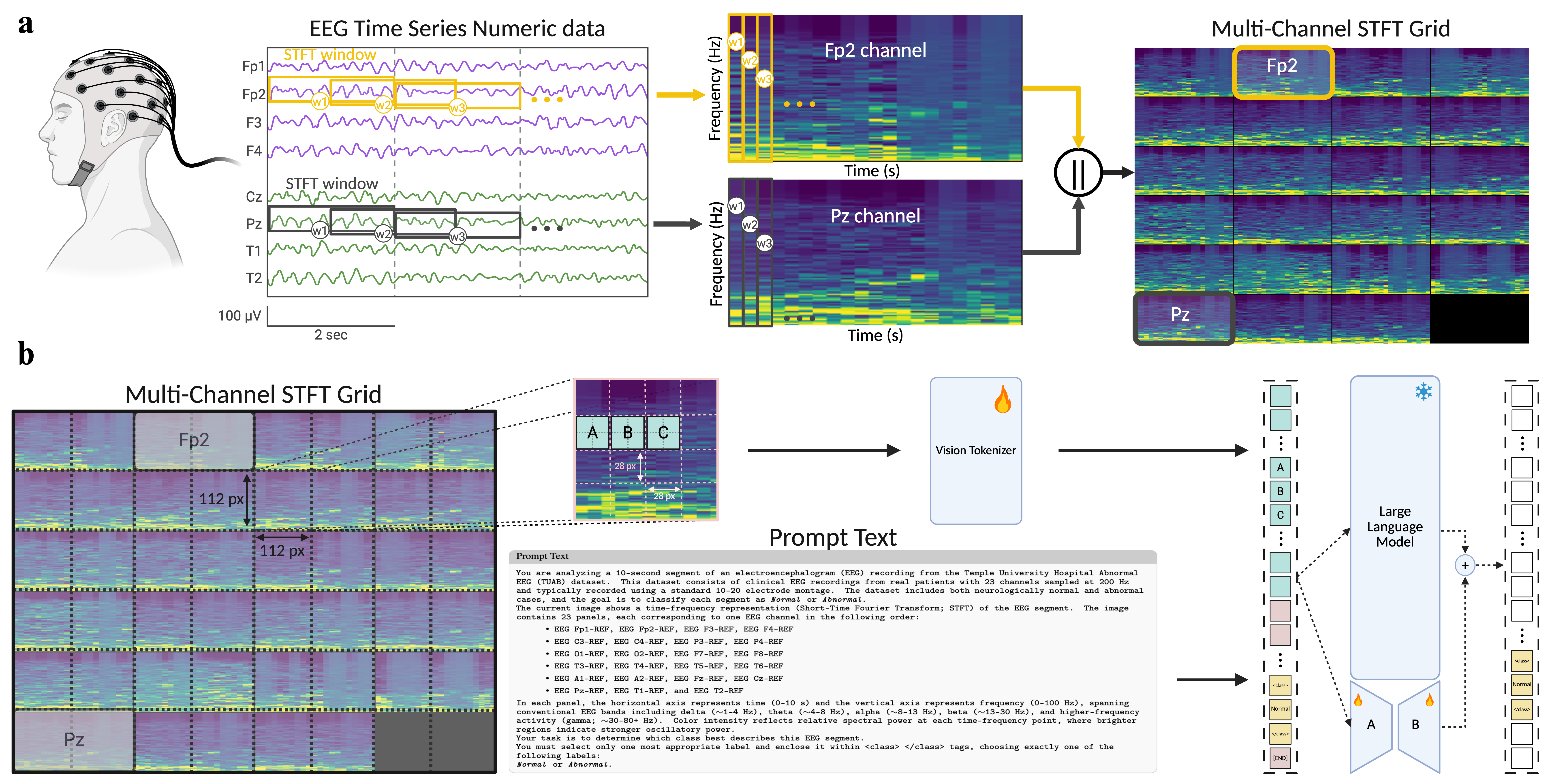}
\caption{\textbf{BraVista framework.} (a) Per-channel STFT spectrograms are arranged into a structured EEG image. (b) A fine-tuned vision encoder processes the image, while a task instruction specifies the dataset, decoding objective, and valid labels. Visual and textual tokens jointly condition a language decoder adapted with LoRA.}

\label{fig:architecture}
\vspace{-0.12in}
\end{figure*}

\section{Method}
The key idea of \ourmethod{} is to use visual representations as an interface that aligns EEG signals with the inductive biases of pretrained vision–language models. Given a multichannel EEG signal, we first transform it into a structured visual representation using a time–frequency mapping. This representation is paired with a task-specific textual instruction, and both modalities are processed jointly by a pretrained VLM.
\vspace{-5pt}
\paragraph{EEG Time–Frequency Representation.}
We begin by formalizing the input EEG signals as \(\mathbf{X} \in \mathbb{R}^{C \times T}\), where $C$ denotes the number of electrodes and $T$ the number of time samples. The signal is partitioned into segments of length $L$, and then each segment is transformed into a time–frequency representation using the STFT, producing a spectrogram for each channel (see Appendix~\ref{appendix:stft_setting} for STFT parameters).
Then, to form a unified visual input, we arrange channel-wise spectrograms into a structured grid (Fig.~\ref{fig:architecture}a). 
This results in an image-like representation that preserves discriminative time–frequency structure and provides a two-dimensional representation that allows vision-based models to capture temporal–spectral correlations~\cite{li2018hierarchical,li2022deep}. This representation is well suited for EEG analysis~\cite{schirrmeister2017deep,craik2019deep,wang2023brainbert}, as EEG signals exhibit non-stationary temporal–spectral dynamics, with task-relevant neural patterns encoded jointly across time and frequency~\cite{makeig2002dynamic}.

\vspace{-5pt}
\paragraph{EEG Instruction Construction.}
For each EEG sample, we construct structured instruction–response pairs that describe the decoding task in natural language. Our instructions provide contextual information about the EEG signal, conclude with an explicit task specification, and the response is defined by the corresponding ground-truth label. This design enables casting EEG decoding as an instruction-following problem and aligns with the training paradigm of VLMs, allowing the model to leverage its pretrained language understanding to perform instruction-conditioned prediction over EEG representations (see Appendix~\ref{app:results} for task-specific instruction details).
\vspace{-5pt}
\paragraph{Training Framework of \ourmethod{}.}
\label{method:sft}
To adapt a general-purpose VLM to EEG data, we perform continued post-training on the \texttt{Qwen3-VL-2B} model~\cite{bai2025qwen3}, whose pretrained visual understanding provides the foundation for EEG domain adaptation. We keep the original model architecture, which consists of three components: a SigLIP-2 vision encoder~\cite{tschannen2025siglip}, a vision–language merger~\cite{liu2023visual}, and a language model decoder. We fully fine-tune the vision encoder to capture task-relevant temporal–spectral patterns in EEG data. This allows the model to align its visual representations with the structure of EEG signals while leveraging pretrained visual priors. To adapt the language model while preserving its pretrained linguistic and generative capabilities, we employ Low-Rank Adaptation (LoRA)~\cite{hu2022lora}, targeting the attention and MLP layers of the decoder. This formulation constrains task-specific updates to a low-dimensional subspace, enabling parameter-efficient adaptation while limiting drift from the pretrained language model.

\paragraph{Optimization for Multi-task EEG Decoding.}
We formulate multi-task EEG decoding as a conditional generation problem over multimodal inputs. Each input consists of an EEG image $\mathcal{I}_{\mathrm{EEG}}$ and a textual instruction $\mathbf{q} = \{q_1, \ldots, q_M\}$. Given these multimodal inputs, we define \textbf{\ourmethod{}} as \(\pi_\theta\) parameterized by $\theta$ as an autoregressive conditional distribution over the output sequence $\mathbf{y}=\{y_1,\ldots,y_K\}$:
\begin{equation}
\label{eq_conditional}
\pi_{\theta}(\mathbf{y} \mid \mathcal{I}_{\mathrm{EEG}}, \mathbf{q})
= \prod_{k=1}^{K} \pi_{\theta}(y_k \mid y_{<k}, \mathcal{I}_{\mathrm{EEG}}, \mathbf{q}),
\end{equation}
where $K$ is the length of the generated output sequence. This formulation allows the model to condition its predictions jointly on EEG visual features, task-level textual context, and prior outputs. To learn the conditional distribution in Eq.~(\ref{eq_conditional}), we use the supervised fine-tuning (SFT) objective:
\begin{equation}
\mathcal{L}
= - \sum_{k=1}^{K} \log p_{\theta}(y_k \mid y_{<k}, \mathcal{I}_{\mathrm{EEG}}, \mathbf{q}).
\end{equation}
This objective trains the model to generate task-specific outputs conditioned jointly on the EEG visual input and the textual prompt, consistent with instruction-following training in LLMs~\cite{ouyang2022training, touvron2023llama}.
During training, gradients are propagated through the vision encoder and the LoRA parameters of the language model, enabling coordinated adaptation of visual and linguistic representations for multi-task EEG decoding. Implementation details are provided in Appendix~\ref{app:hyperparams}.

\section{Experiments and Results}
\subsection{Evaluation Datasets and Baselines}
We evaluate one jointly trained model on HMC sleep staging~\cite{alvarez2021inter}, SEED emotion recognition~\cite{zheng2015investigating}, EEGMAT cognitive-workload classification~\cite{zyma2019electroencephalograms}, and TUAB abnormality detection~\cite{obeid2016temple}. Dataset statistics are summarized in Table~\ref{tab:datasets}. We follow prior preprocessing and splits~\cite{jiang2024large,jiang2024neurolm,zhou2025csbrain}. Furthermore, for the SEED dataset, we explicitly exclude foundation models whose pretraining data includes SEED to prevent data leakage. We report balanced accuracy as the primary evaluation metric and additionally report weighted F1 scores for ablation studies. All results are presented as mean $\pm$ standard deviation over three runs with different random seeds.

Our comparisons include: (1) supervised baselines trained separately for each dataset; (2) EEG foundation models, in which a single pretrained encoder is fine-tuned per downstream task; and (3) the EEG-language model NeuroLM, which is the closest unified baseline because it combines a learned EEG tokenizer with an instruction-tuned language model. The unadapted Qwen3-VL backbone provides a zero-shot control for whether generic VLMs alone can decode EEG.

\begin{figure*}[t]
  \centering
  \includegraphics[width=\textwidth]{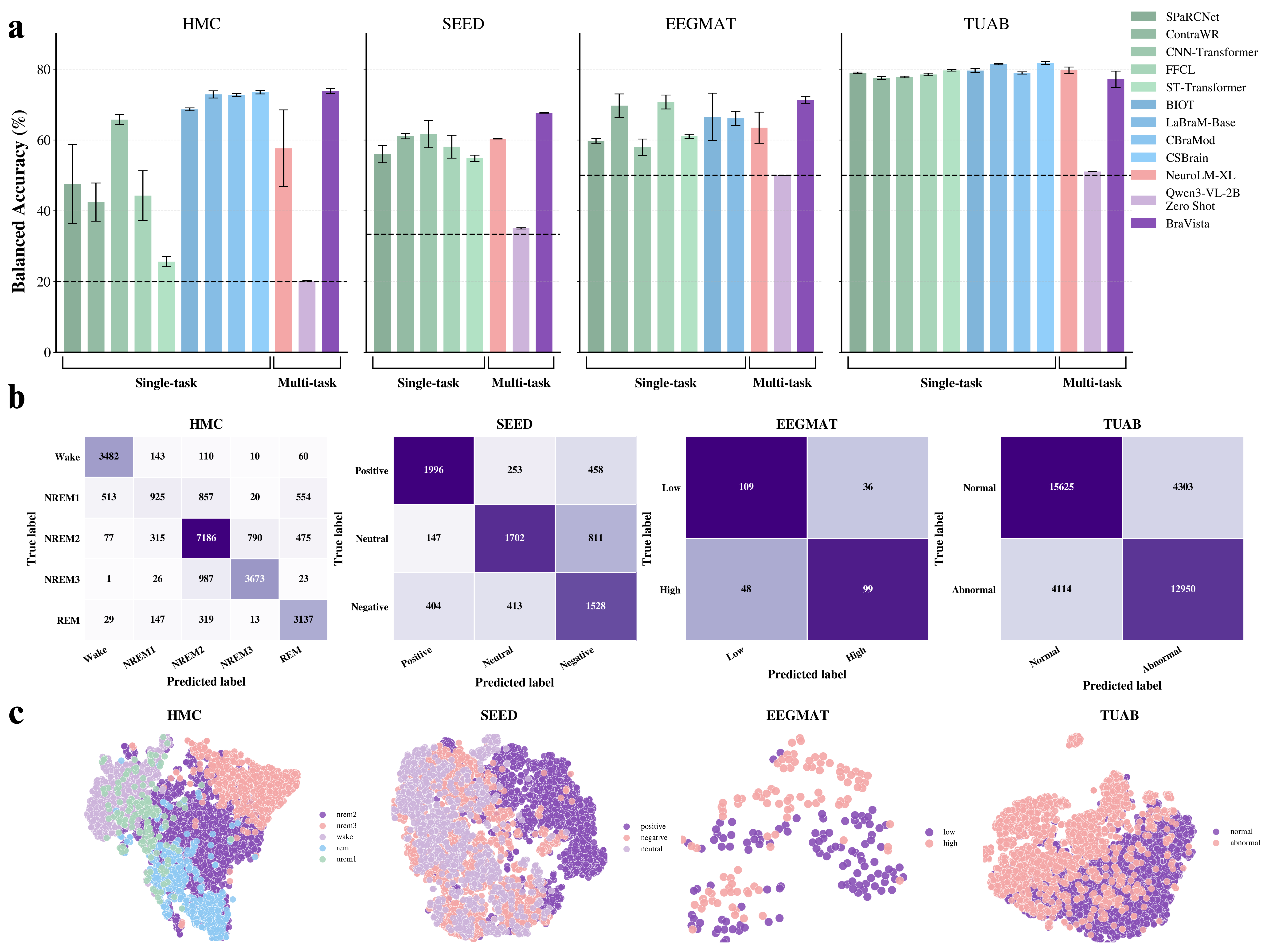}
  \caption{\textbf{Unified decoding across four EEG tasks.} (a) Balanced accuracy of BraVista against supervised models, EEG foundation models, NeuroLM, and the zero-shot VLM backbone. (b) Confusion matrices. (c) t-SNE projections of learned embeddings, colored by class.}
  \label{fig:multitask_performance}
  \vspace{-0.12in}
\end{figure*}

\subsection{\ourmethod{} Enables Multi-Task EEG Decoding Across Diverse BCI Tasks}
We compare \ourmethod{} with task-specific supervised and foundation models, the multi-task NeuroLM baseline~\cite{jiang2024neurolm}, and the zero-shot VLM. As shown in Fig.~\ref{fig:multitask_performance}a, \ourmethod{} consistently achieves strong performance while operating as a single unified model on all datasets.
Importantly, \ourmethod{} substantially surpasses the zero-shot \texttt{Qwen3-VL-2B} across all datasets, highlighting that our post-training effectively aligns visual-linguistic priors with neural signal characteristics.

We further analyze the class-level behavior of \ourmethod{} using confusion matrices (Fig.~\ref{fig:multitask_performance}b).
The observed confusion patterns are consistent with known task-specific challenges. For example, in HMC, the model shows the largest confusion between NREM1 and NREM2, whose distinguishing transient graphoelements (such as sleep spindles and K-complexes) may be absent from a single 30-second epoch~\cite{malhotra2024aasm}. Similarly, for the SEED dataset, the confusion between Neutral and Negative classes reflects their overlapping neural patterns in the beta and gamma bands~\cite{zheng2015investigating}. 
This suggests that many errors reflect intrinsic task difficulty rather than model failure, and that \ourmethod{} captures meaningful physiological structure in EEG signals.
Finally, to evaluate the quality of the learned representations, we visualize the embeddings of \ourmethod{} using t-SNE (Fig.~\ref{fig:multitask_performance}c). The embeddings show clustering patterns corresponding to class labels across datasets.

\subsection{Time--Frequency Images Are the Effective Interface}
\begin{figure}[t]
  \centering
  \includegraphics[width=0.75\textwidth]{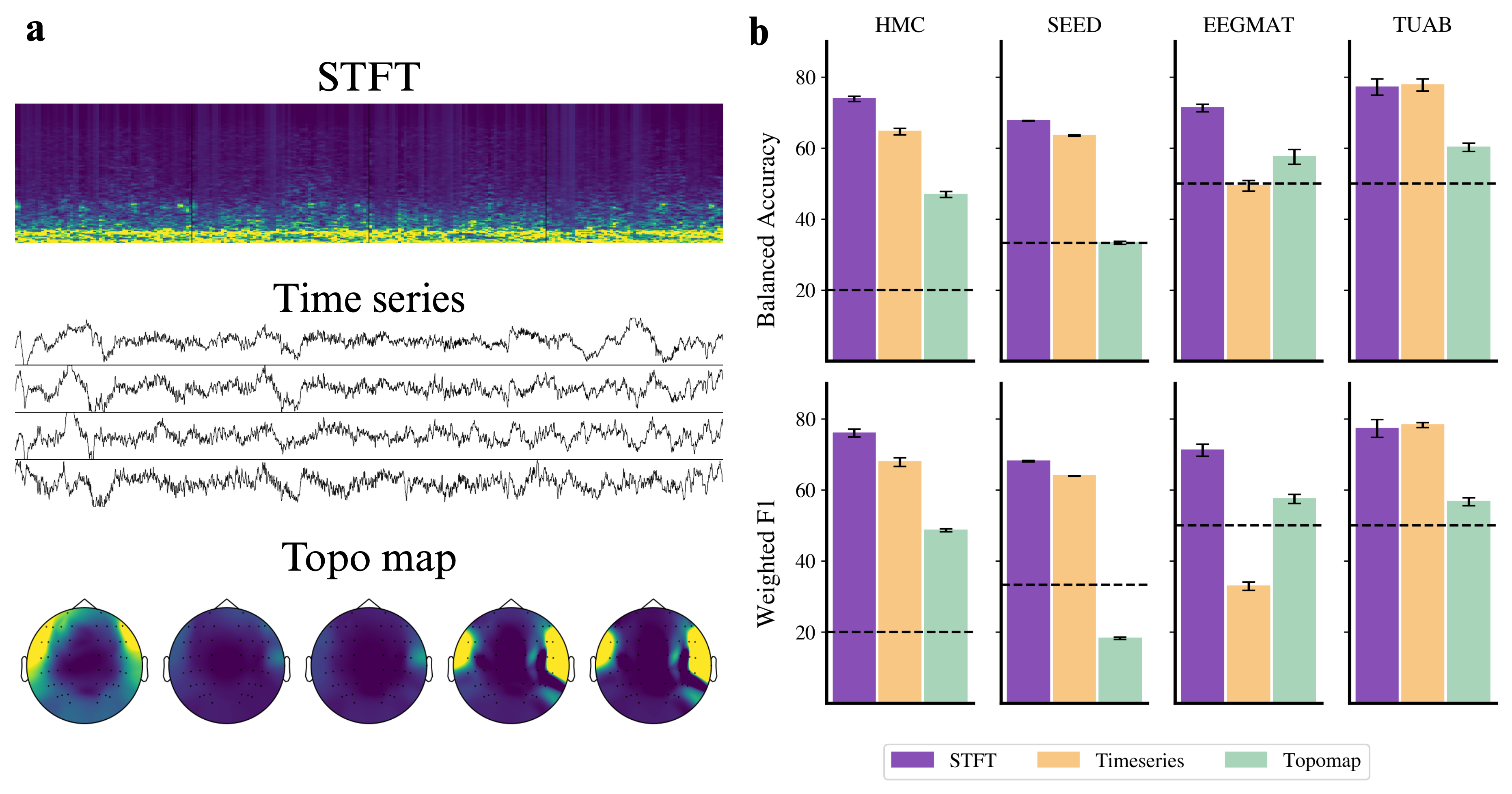}
  \caption{\textbf{The visual interface matters.} (a) STFT, raw time-series, and scalp-topography inputs. (b) Balanced accuracy (top) and weighted F1 (bottom) across four datasets.}
  \label{fig:image_modality}
  \vspace{-0.1in}
\end{figure}
We study how the choice of visual representation affects the ability of VLMs to interpret EEG signals. We replace the STFT grid with raw time-series plots or scalp topographies while holding the model and training procedure fixed. STFT  obtains the highest balanced accuracy and weighted F1 across all four datasets (Fig.~\ref{fig:image_modality}). This result highlights the importance of preserving time–frequency structure when adapting vision–language models to EEG data. Time-series plots retain temporal information and topographic maps capture spatial distributions, but neither provides a joint representation of temporal and spectral dynamics. In contrast, STFT representations provide a dense, structured visualization of neural activity that is more compatible with the visual processing capabilities of pretrained vision encoders.

\subsection{\ourmethod{} Relies on EEG Signal Structure Under Noise Corruption}
\begin{figure}[!ht]
    \centering
    \includegraphics[width=0.5\textwidth]{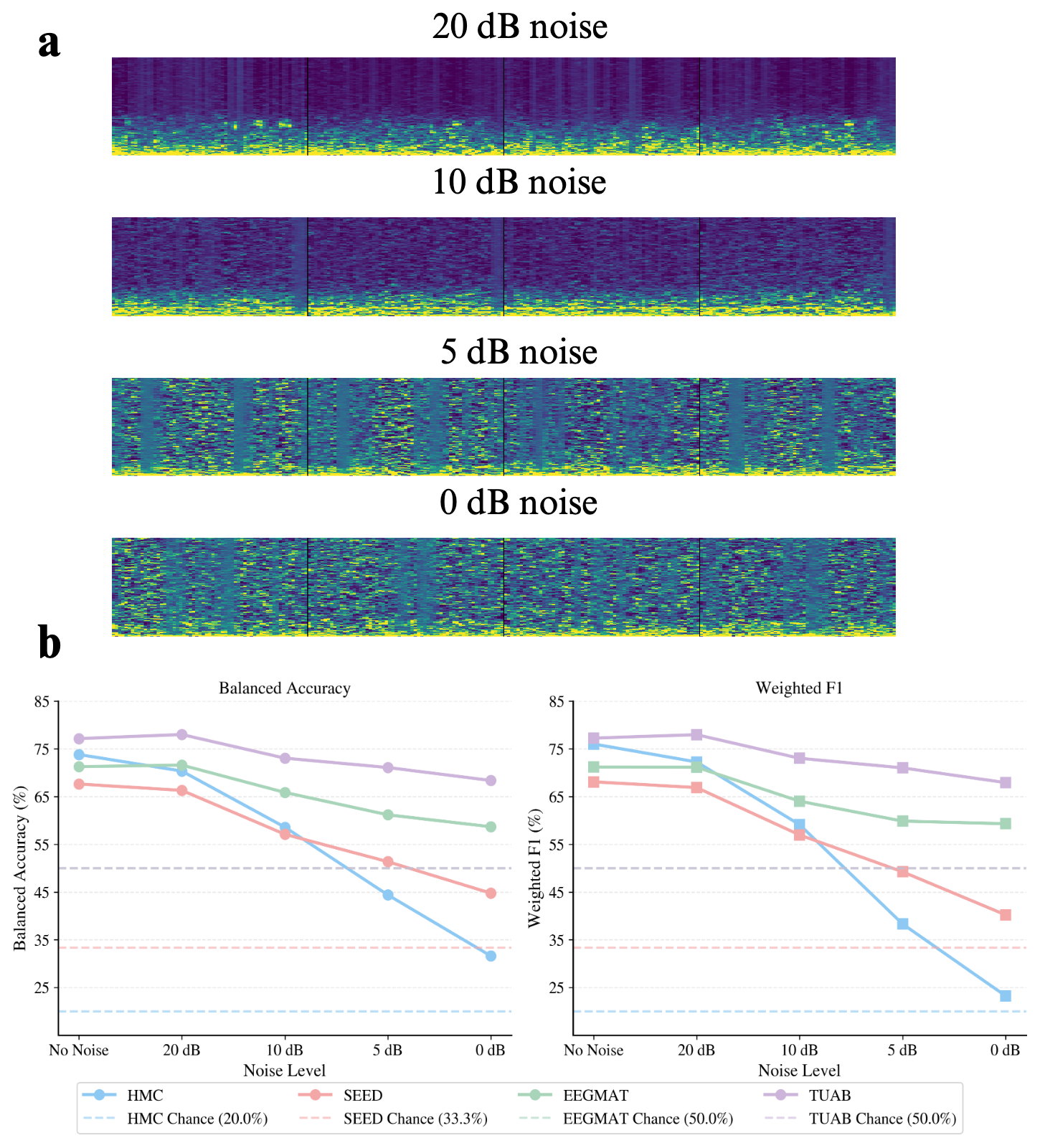}
    \caption{\textbf{Robustness of \ourmethod{} to EEG noise.} \textbf{(a)} Examples of STFT under increasing noise levels (20~dB to 0~dB). \textbf{(b)} Balanced accuracy and weighted F1. Dashed lines indicate chance performance.}
    \label{fig:noise_robustness}
    \vspace{-0.1in}
\end{figure}

To assess whether \ourmethod{} relies on EEG information encoded in the visual representation, we evaluate its behavior under controlled noise corruption by progressively adding noise to the input signals at different signal-to-noise ratios (20~dB, 10~dB, 5~dB, and 0~dB). As noise increases, performance degrades consistently and gradually rather than abruptly across all datasets for both balanced accuracy and weighted F1, and the performance remains above chance even under severe noise conditions (Fig.~\ref{fig:noise_robustness}b). 
This gradual performance decay suggests that predictions depend on the quality of the underlying EEG signal rather than on superficial visual patterns.

\section{Discussion}
In this work, we present \ourmethod{}, a framework that reformulates EEG decoding by interfacing neural signals with general-domain vision–language models. By transforming multichannel EEG recordings into time–frequency images and pairing them with task-specific instructions, we enable unified multi-task decoding within a single model, without task-specific architectures or a separate large-scale EEG-specific pretraining stage. This framework shifts the focus from designing specialized EEG models to designing effective interfaces between neural signals and general-domain foundation models. Visual representations offer one such interface, enabling the reuse of powerful vision–language models for neural signal decoding.

Our results provide two main insights. First, the choice of EEG-to-image representation strongly affects decoding performance, with STFT spectrograms providing the most effective interface among the representations evaluated. Second, the gradual degradation under noise corruption, physiologically plausible class-level errors, and structure of the learned embeddings provide convergent evidence that \ourmethod{} uses EEG-relevant information rather than relying only on superficial visual patterns.

Despite these promising results, several limitations point to important directions for future work. While our framework enables instruction-conditioned multi-task decoding, it currently focuses on producing class labels without explicitly leveraging the reasoning capabilities of VLMs. Integrating reinforcement learning (RL) strategies, such as fine-tuning with process- or outcome-based reward models, presents a promising direction to explicitly guide model reasoning across temporal EEG segments, minimize false positives, and optimize step-by-step diagnostic hypotheses. Additionally, enabling models to reason or integrate contextual information (e.g., subject-specific metadata or clinical history) may improve performance on more complex and clinically relevant tasks~\cite{torabi2026neuromamballm}. Exploring such capabilities would move beyond classification toward more interpretable and interactive EEG analysis, better aligning neural decoding with the broader strengths of large multimodal models. Beyond decoding, a unified neural decoder could serve as the readout component of closed-loop neurotechnologies that pair neural recording with electrical stimulation~\cite{sharafi2026population,householder2026extraocular,sharafi2026high}. Moreover, because \ourmethod{} operates on standard image inputs, it may integrate naturally with connected healthcare and medical imaging infrastructure~\cite{einollahi2023internet}. Overall, this work suggests that representation design, combined with general-domain foundation models, provides a promising path toward scalable and unified EEG decoding.

\begin{ack}
This work was supported by National Institutes of Health (NIH) grants R61MH135407, R01MH123770, and RF1DA056402, the One Mind Rising Star Award, and the Foundation for OCD Research (FFOR).
\end{ack}
\bibliography{example_paper}
\bibliographystyle{unsrt}
\appendix
\onecolumn
\section*{Appendix}
\section{Time--Frequency Representation Parameters}
\label{appendix:stft_setting}
For each channel $c \in C$, the STFT is computed as
\begin{equation}
X_c(\tau, \omega) = \sum_{t} x_c(t)\, w(t-\tau)\, e^{-j\omega t},
\end{equation}
where $w(\cdot)$ denotes a window function and $\tau$ indexes the window center. We use a window size of 400 samples (2 seconds) with an overlap of 300 samples (1.5 seconds).
We retain the magnitude spectrogram
\begin{equation}
S_c(\tau, f) = \left| X_c(\tau, 2\pi f) \right|,
\end{equation}
which captures localized spectral energy over time. The frequency axis is also truncated to a fixed upper bound of 100 Hz to focus on physiologically relevant EEG bands and to suppress high-frequency noise. 

\section{Baselines}
Here, we categorize the baseline methods used for our comparisons:
\paragraph{Task-specific supervised models:}
\begin{itemize}
    \item \textbf{SPaRCNet}~\cite{jing2023development} is a deep neural network based on one-dimensional CNNs with dense residual connections for EEG decoding.
    \item \textbf{ContraWR}~\cite{yang2021self} is a CNN-based model that first transforms EEG signals into multi-channel spectrogram representations and then applies a ResNet-style two-dimensional CNN to extract discriminative features from the spectrograms.
    \item \textbf{CNN-Transformer}~\cite{peh2022transformer} utilizes CNNs to extract local features and transformer layers to model global dependencies in EEG signals.
    \item \textbf{FFCL}~\cite{li2022motor} employs parallel CNN and LSTM layers, where the CNN captures spatial features and the LSTM models temporal dynamics; features from both layers are then fused for classification.
    \item \textbf{ST-Transformer}~\cite{song2021transformer} is a transformer-based architecture that uses attention mechanisms to learn spatial and temporal representations from EEG signals.
\end{itemize}

\paragraph{Foundation models:}
\begin{itemize}
    \item \textbf{BIOT}~\cite{yang2023biot} is a transformer-based model for learning general-purpose biosignal representations using a mix of supervised and unsupervised pretraining. The pretrained model supports up to 18 EEG channels as input. For recordings with more than 18 channels, a $1 \times 1$ convolution is used to reduce the input to 18 channels.
    \item \textbf{LaBraM}~\cite{jiang2024large} is a large-scale EEG foundation model pretrained in an unsupervised manner on over 2,500 hours of diverse EEG data. It learns universal EEG representations by segmenting raw signals into channel-wise patches and employing vector-quantized neural spectrum prediction as a tokenizer, with a Transformer backbone capturing both temporal and spatial features for downstream applicability.
    \item \textbf{CBraMod}~\cite{wang2024cbramod} is an EEG foundation model that learns generic representations via patch-based masked EEG reconstruction. It employs a criss-cross transformer to jointly model spatial and temporal dependencies across EEG patches in parallel, along with an asymmetric convolutional positional encoding to capture diverse spatial–temporal signal formats.
    \item \textbf{CSBrain}~\cite{zhou2025csbrain} is a cross-subject brain foundation model that captures shared neural patterns across individuals through large-scale pretraining, aiming to enhance robustness and adaptability to downstream EEG analysis tasks.

\end{itemize}

\paragraph{Multi-task LLMs:}
\begin{itemize}
    \item \textbf{NeuroLM}~\cite{jiang2024neurolm} is a multi-task EEG foundation model that learns a neural tokenizer that converts EEG signals into discrete tokens, which are passed to a large language model. It is pretrained on a large-scale EEG corpus to learn generic representations and is fine-tuned using instruction tuning.
\end{itemize}

\section{Experimental Settings}

\subsection{Implementation Details}
We train \ourmethod{} for 1 epoch with an effective batch size of 128, using a local batch size of 8 and gradient accumulation steps of 4. Training is performed on 4 NVIDIA RTX A6000 GPUs for 12 hours. We apply a linear warm-up over the first 3\% of total training steps, followed by a cosine learning rate decay schedule. The language model decoder is optimized using the AdamW optimizer with a peak learning rate of $1 \times 10^{-4}$, while the vision encoder is trained with a learning rate of $2 \times 10^{-6}$.

\subsection{Hyperparameter Settings}\label{app:hyperparams}

\vspace{5pt}
\begin{table}[!ht]
\centering
\caption{Post-training hyperparameters.}
\label{tab:hyperparams}
\begin{tabular}{l l}
\hline
\textbf{Hyperparameters} & \textbf{Value} \\
\hline
Base Model & Qwen3-VL-2B-Instruct \\
Training mode & LLM LoRA + vision encoder \\
Batch size per device & 8 \\
Number of devices & 4 \\
Global batch size & 128 \\
Gradient accumulation steps & 4 \\
Epochs & 1 \\
Learning rate (LLM/LoRA parameters) & $1 \times 10^{-4}$ \\
Learning rate (Vision) & $2 \times 10^{-6}$ \\
Weight decay & 0.1 \\
Warmup ratio & 0.03 \\
LR scheduler & Cosine \\
Freeze vision tower & False \\
Freeze LLM & True \\
Tune merger & False \\
LoRA rank ($r$) & 32 \\
LoRA scaling factor ($\alpha$) & 64 \\
LoRA dropout & 0.05 \\
\hline
\vspace{1pt}
\end{tabular}
\end{table}

The hyperparameter settings in Table~\ref{tab:hyperparams} correspond to the jointly trained \ourmethod{} model.

\subsection{Metrics}\label{app:metrics}
In this section, we describe the evaluation metrics used throughout the paper. We report evaluation metrics that account for class imbalance and task heterogeneity, which are common characteristics of EEG datasets. In many EEG classification settings, label distributions are imbalanced across classes, making standard accuracy insufficient to fully capture model performance. Following prior work in EEG foundation models~\cite{jiang2024large}, we adopt metrics that provide a more balanced and informative evaluation.

\begin{itemize}
\item \textbf{Balanced Accuracy} is computed as the average recall across all classes. It accounts for imbalanced label distributions by giving equal weight to each class and is used for both binary and multi-class classification tasks. This metric is particularly suitable for EEG classification tasks with imbalanced label distributions, as it prevents performance on majority classes from dominating the overall score.

\item \textbf{Weighted F1} is the average of per-class F1 scores, weighted by the number of samples in each class. It captures both precision and recall while reflecting the overall data distribution, providing a complementary view of performance that emphasizes effectiveness on more prevalent classes.

\end{itemize}

\section{Dataset Details}\label{app:results}
This section describes the datasets, preprocessing procedures, data splitting protocols, and the generated image and instruction for each dataset.

\begin{table}[!h]
\centering
\small
\caption{Evaluation datasets span heterogeneous acquisition settings and tasks.}
\label{tab:datasets}
\begin{adjustbox}{width=\textwidth}
\begin{tabular}{llrrrr}
\toprule
Dataset & Task & Channels & Sampling & Segment & Samples \\
\midrule
HMC & Sleep stage (5-class) & 4 & 256 Hz & 30 s & 137,243 \\
SEED & Emotion (3-class) & 62 & 1000 Hz & 4 s & 38,475 \\
EEGMAT & Workload (binary) & 19 & 500 Hz & 4 s & 2,088 \\
TUAB & Abnormality (binary) & 23 & 256 Hz & 10 s & 409,455 \\
\bottomrule
\end{tabular}
\end{adjustbox}
\vspace{-0.08in}
\end{table}
\subsection{Sleep Stage Classification}
The Haaglanden Medisch Centrum (HMC)~\cite{alvarez2021inter} sleep staging dataset is a large EEG corpus collected for automated sleep stage classification. It contains overnight EEG recordings from 151 subjects with segments labeled according to five standard sleep stages: \textit{Wake}, \textit{NREM1}, \textit{NREM2}, \textit{NREM3}, and \textit{REM}. EEG signals were recorded from four channels (F4–M1, C4–M1, O2–M1, and C3–M2) using a clinical 10–20 montage and sampled at 256Hz. Following the NeuroLM~\cite{jiang2024neurolm} preprocessing settings, the raw signals are bandpass filtered between 0.1Hz and 75Hz, notch filtered at 50Hz to mitigate power-line interference, resampled to 200Hz, and normalized by dividing signal amplitudes by 100. For data division, the first 100 subjects are used for training, the next 25 subjects for validation, and the final 26 subjects for testing.

\begin{figure*}[!ht]
  \vskip 0.2in
  \begin{center}
    \centerline{\includegraphics[width=0.5\textwidth]{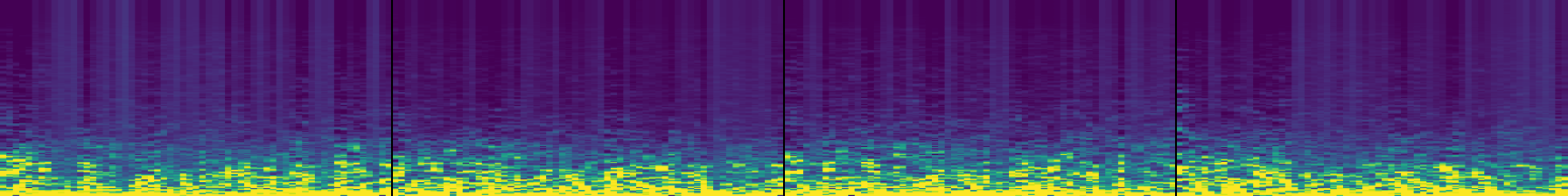}}
    \caption{
    Sample HMC STFT image.}
    \label{fig:hmc_stft}
  \end{center}
  \vskip -0.2in
\end{figure*}

Each preprocessed EEG segment is converted into an STFT image and paired with a dataset-specific natural language prompt for sleep stage classification. An example STFT image for the sleep staging dataset is shown in Figure~\ref{fig:hmc_stft}. The STFT representations of the EEG channels are arranged in a grid according to the channel ordering provided by the dataset. In particular, the 4 channels are placed into a \(1\times4\) grid, where each channel has a resolution of \(224\times448\) resulting in a final image resolution of \(224\times1792\).

\begin{eegresponse}[EEG Sleep Stage Classification Prompt]
<image> You are analyzing a 30-second segment from a full-night electroencephalogram (EEG) recording from the Haaglanden Medisch Centrum (HMC) sleep staging dataset.

This dataset contains EEG recordings from 151 subjects, with segments corresponding to one of the following 5 sleep stages:

Wake (Conscious awareness), NREM1 (Sleep onset), NREM2 (Light stable sleep), NREM3 (Deep restorative sleep), and REM (Dream associated sleep).

Recordings include EEG sampled at 200 Hz recorded from 4 channels: EEG F4-M1, EEG C4-M1, EEG O2-M1, and EEG C3-M2.

The current image shows a time-frequency representation (Short-Time Fourier Transform; STFT) of the EEG segment.

The image contains four panels (left to right): EEG F4-M1, EEG C4-M1, EEG O2-M1, and EEG C3-M2.

In each panel, the horizontal axis represents time (0-30 s) and the vertical axis represents frequency (0-100Hz), spanning conventional EEG bands including delta ($\sim$1-4 Hz), theta ($\sim$4-8 Hz), alpha ($\sim$8-13 Hz), beta ($\sim$13-30 Hz), and higher-frequency activity (gamma; $\sim$30-80+ Hz).

Color intensity reflects relative spectral power at each time–frequency point, where brighter regions indicate stronger oscillatory power.

Your task is to determine the sleep stage associated with this 30-second EEG segment.

You MUST only select ONE most appropriate label and enclose it within <class> </class> tags, choosing exactly one of the following labels:

"Wake", "NREM1", "NREM2", "NREM3", or "REM".
\end{eegresponse}

\subsection{Emotion Recognition}
SEED (SJTU Emotion EEG Dataset)~\cite{zheng2015investigating} comprises EEG recordings from 15 subjects watching emotion-eliciting Chinese film clips corresponding to three affective states: \textit{positive}, \textit{neutral}, and \textit{negative}. EEG signals were recorded using 62 channels at a sampling rate of 1000 Hz across three experimental sessions per subject, with each session containing 15 trials evenly distributed across the three emotion categories. Following the NeuroLM~\cite{jiang2024neurolm} preprocessing settings, the EEG signals are bandpass filtered between 0.1 Hz and 75 Hz, notch filtered at 50 Hz to remove power-line interference, resampled to 200 Hz, and normalized by dividing signal amplitudes by 100. For data splitting, we follow a chronological trial-based protocol: the 15 trials are split into training, validation, and test sets using a 9:3:3 ratio, respectively, and all sessions are merged to form the final splits.

\begin{figure*}[!ht]
  \vskip 0.2in
  \begin{center}
    \centerline{\includegraphics[width=0.5\textwidth]{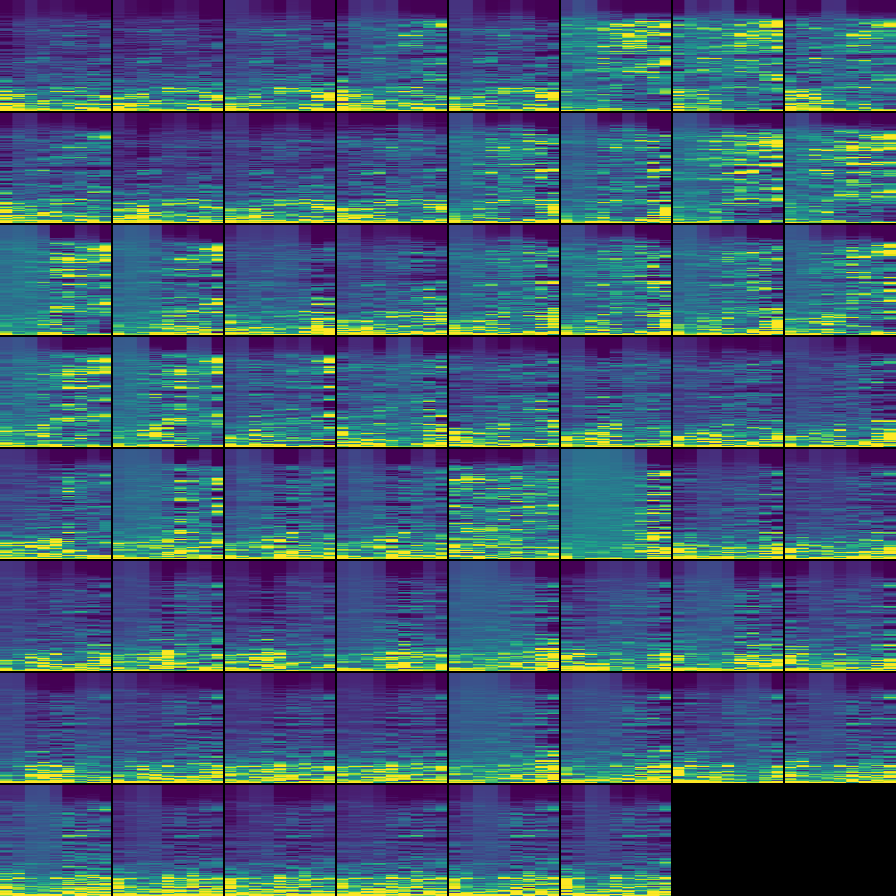}}
    \caption{
    Sample SEED STFT image.}
    \label{fig:seed_stft}
  \end{center}
  \vskip -0.2in
\end{figure*}

Each preprocessed EEG segment is converted into an STFT image and paired with a dataset-specific natural language prompt for emotion classification. An example STFT image for the SEED dataset is shown in Figure~\ref{fig:seed_stft}. The STFT representations of the EEG channels are arranged in a grid according to the channel ordering provided by the dataset. In particular, the 62 channels are placed into an \(8\times8\) grid, where each channel has a resolution of \(112\times112\) resulting in a final image resolution of \(896\times896\).

\begin{eegresponse}[EEG Emotion Recognition Prompt]
<image> You are analyzing a 4-second segment from an electroencephalogram (EEG) recording from the SJTU Emotion EEG Dataset (SEED).

This dataset contains EEG recordings collected from multiple subjects while they watched emotionally evocative video clips.

Each EEG segment corresponds to one of the following emotional states:

Positive (pleasant or high-valence emotion), Neutral (emotionally calm or baseline), or Negative (unpleasant or low-valence emotion).

Recordings include EEG signals sampled at 200 Hz and recorded from 62 channels.

The current image shows a time–frequency representation (Short-Time Fourier Transform; STFT) of the EEG segment.

The image contains 62 panels corresponding to each of the EEG channels.

In each panel, the horizontal axis represents time (seconds) and the vertical axis represents frequency (0-100 Hz), spanning standard EEG bands including delta ($\sim$1-4 Hz), theta ($\sim$4-8 Hz), alpha ($\sim$8-13 Hz), beta ($\sim$13-30 Hz), and higher-frequency activity (gamma; $\sim$30-80+ Hz).

Color intensity reflects relative spectral power at each time–frequency point, where brighter regions indicate stronger oscillatory power.

Your task is to determine the emotional state associated with this EEG segment.

You MUST only select ONE most appropriate label and enclose it within <class> </class> tags, choosing exactly one of the following labels:

"Positive", "Neutral", or "Negative".
\end{eegresponse}

\subsection{Mental Arithmetic Task}
The EEG During Mental Arithmetic Tasks dataset (EEGMAT)~\cite{zyma2019electroencephalograms} comprises EEG recordings from 36 subjects performing mental arithmetic tasks and resting-state conditions. EEG signals were recorded using 19 channels arranged according to the international 10–20 system and sampled at 500~Hz. During the task condition, subjects performed a serial subtraction task, inducing a high cognitive workload, while the resting condition corresponds to a low workload state. Each recording consists of continuous EEG data, which we segment into non-overlapping 4-second windows for classification. The EEG signals are bandpass filtered between 0.1~Hz and 75~Hz, notch filtered at 50~Hz to remove power-line interference, resampled to 200~Hz, and normalized by dividing signal amplitudes by 100. We adopt a subject-independent split, using subjects 0–25 for training, 26–30 for validation, and 31–35 for testing.

\begin{figure*}[!ht]
  \vskip 0.2in
  \begin{center}
    \centerline{\includegraphics[width=0.5\textwidth]{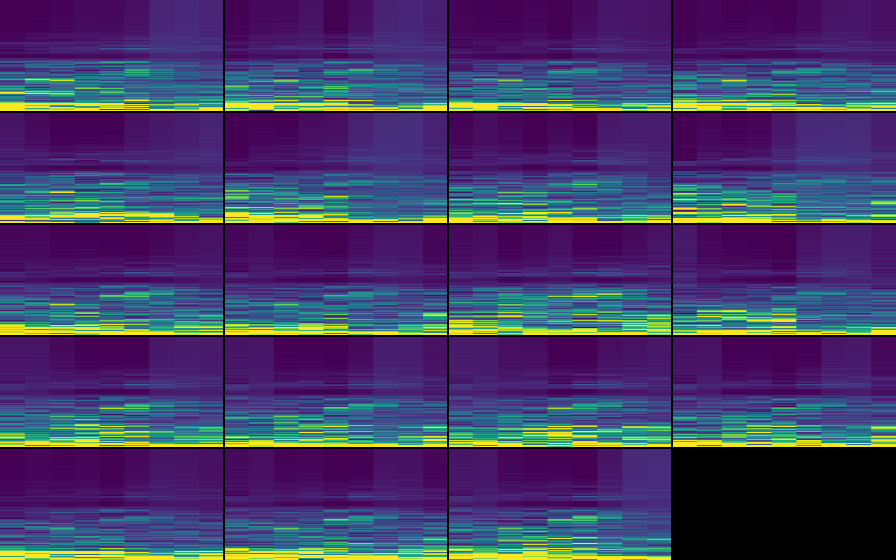}}
    \caption{
    Sample EEGMAT STFT image.}
    \label{fig:workload_stft}
  \end{center}
  \vskip -0.2in
\end{figure*}

Each preprocessed EEG segment is converted into an STFT image and paired with a dataset-specific natural language prompt for mental arithmetic detection. An example STFT image for the mental arithmetic dataset is shown in Figure~\ref{fig:workload_stft}. The STFT representations of the EEG channels are arranged in a grid according to the channel ordering provided by the dataset. In particular, the 19 channels are placed into a \(5\times4\) grid, where each channel has a resolution of \(112\times224\) resulting in a final image resolution of \(560\times896\).

\begin{eegresponse}[EEG Mental Arithmetic Detection Prompt]
<image> You are analyzing a 4-second segment of an electroencephalogram (EEG) recording from the EEG During Mental Arithmetic Tasks (Zyma et al., 2019) dataset.

This dataset contains EEG recordings from 36 subjects from 19 channels sampled at 200 Hz and typically recorded using a standard 10-20 electrode montage.

Each EEG segments corresponds either to serial subtraction tasks (high cognitive workload) or to resting-state EEG (low cognitive workload).

The current image shows a time-frequency representation (Short-Time Fourier Transform; STFT) of the EEG segment.

The image contains 19 panels each corresponding to one EEG channel in the following order:

EEG Fp1, EEG Fp2, EEG F3, EEG F4, 

EEG F7, EEG F8, EEG T3, EEG T4, 

EEG C3, EEG C4, EEG T5, EEG T6, 

EEG P3, EEG P4, EEG O1, EEG O2, 

EEG Fz, EEG Cz, EEG Pz. 

In each panel, the horizontal axis represents time (0-4 s) and the vertical axis represents frequency (0-100Hz), spanning conventional EEG bands including delta ($\sim$1-4 Hz), theta ($\sim$4-8 Hz), alpha ($\sim$8-13 Hz), beta ($\sim$13-30 Hz), and higher-frequency activity (gamma; $\sim$30-80+ Hz).

Color intensity reflects relative spectral power at each time–frequency point, where brighter regions indicate stronger oscillatory power.

Your task is to determine the cognitive workload level associated with this EEG segment.

You MUST only select ONE most appropriate label and enclose it within <class> </class> tags, choosing exactly one of the following labels:
"Low" or "High".
\end{eegresponse}

\subsection{Abnormality Detection}

The Temple University Hospital Abnormal EEG (TUAB) dataset~\cite{obeid2016temple} is a large-scale clinical EEG corpus designed for abnormality detection in EEG recordings. It contains EEG data collected from patients in a clinical setting, where each recording is labeled as either \textit{normal} or \textit{abnormal} based on expert neurological assessment. EEG signals are recorded using 23 channels following the standard 10–20 electrode montage and sampled at 256~Hz. For this task, continuous EEG recordings are segmented into non-overlapping 10-second clips, each inheriting the label of the corresponding recording. The EEG signals are bandpass filtered between 0.1~Hz and 75~Hz, notch filtered at 50~Hz to remove power-line interference, resampled to 200~Hz, and normalized by dividing signal amplitudes by 100. The dataset provides an official patient-level split into training and test sets; we further divide the training set into training and validation subsets using an 80\%:20\% ratio.

\begin{figure*}[!ht]
  \vskip 0.2in
  \begin{center}
    \centerline{\includegraphics[width=0.5\textwidth]{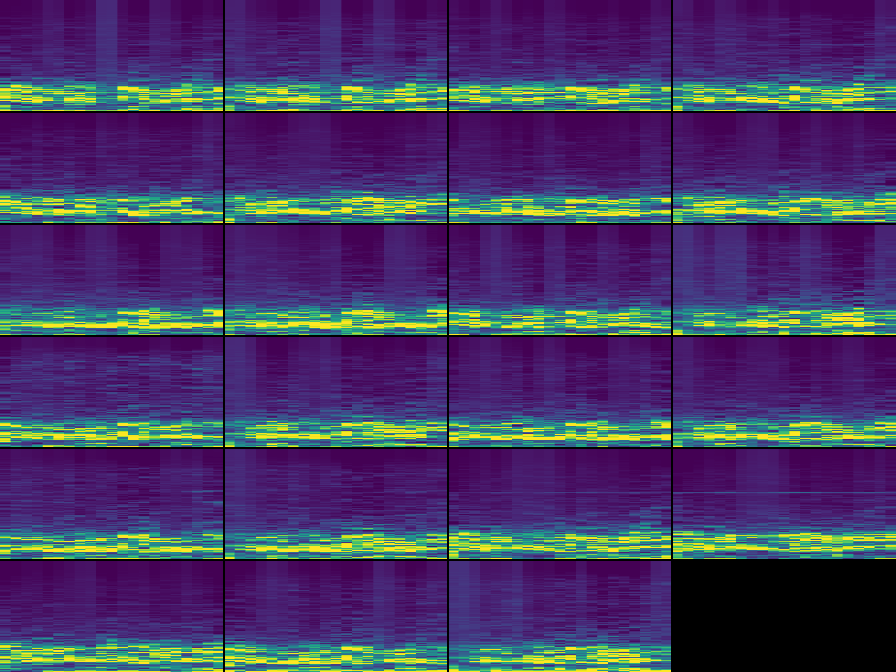}}
    \caption{
    Sample TUAB STFT image.}
    \label{fig:tuab_stft}
  \end{center}
  \vskip -0.2in
\end{figure*}

Each preprocessed EEG segment is converted into an STFT image and paired with a dataset-specific natural language prompt for abnormality detection. An example STFT image for the TUAB dataset is shown in Figure~\ref{fig:tuab_stft}. The STFT representations of the EEG channels are arranged in a grid according to the channel ordering provided by the dataset. In particular, the 23 channels are placed into a \(6\times4\) grid, where each channel has a resolution of \(112\times224\) resulting in a final image resolution of \(672\times896\).
\begin{eegresponse}[EEG Abnormal Detection Prompt]
<image> You are analyzing a 10-second segment of an electroencephalogram (EEG) recording from the Temple University Hospital Abnormal EEG (TUAB) dataset.

This dataset consists of clinical EEG recordings from real patients from 23 channels sampled at 200 Hz and typically recorded using a standard 10-20 electrode montage.

The dataset includes both neurologically normal and abnormal cases, and the goal is to classify each segment as "Normal" or "Abnormal".

The current image shows a time–frequency representation (Short-Time Fourier Transform; STFT) of the EEG segment.

The image contains 23 panels each corresponding to one EEG channel in the following order:

EEG Fp1-REF, EEG Fp2-REF, EEG F3-REF, EEG F4-REF, 

EEG C3-REF, EEG C4-REF, EEG P3-REF, EEG P4-REF, 

EEG O1-REF, EEG O2-REF, EEG F7-REF, EEG F8-REF, 

EEG T3-REF, EEG T4-REF, EEG T5-REF, EEG T6-REF, 

EEG A1-REF, EEG A2-REF, EEG Fz-REF, EEG Cz-REF, 

EEG Pz-REF, EEG T1-REF, and EEG T2-REF. 

In each panel, the horizontal axis represents time (0-10 s) and the vertical axis represents frequency (0-100Hz), spanning conventional EEG bands including delta ($\sim$1-4 Hz), theta ($\sim$4-8 Hz), alpha ($\sim$8-13 Hz), beta ($\sim$13-30 Hz), and higher-frequency activity (gamma; $\sim$30-80+ Hz).

Color intensity reflects relative spectral power at each time–frequency point, where brighter regions indicate stronger oscillatory power.

Your task is to determine which class best describes this EEG segment.

You MUST only select ONE most appropriate label and enclose it within <class> </class> tags, choosing exactly one of the following labels:

"Normal" or "Abnormal".
\end{eegresponse}

\end{document}